\documentclass[11pt]{article}

\usepackage[utf8]{inputenc}
\usepackage[T1]{fontenc}
\usepackage{lmodern}
\usepackage{geometry}
\usepackage{microtype}
\usepackage{enumitem}
\usepackage{amsmath}
\usepackage{url}
\usepackage[colorlinks=true,allcolors=blue]{hyperref}

\setlist{nosep}
\title{Agentic AI Translate: An Agentic Translator Prototype for Translation as Communication Design}
\author{Masaru Yamada\\Rikkyo University; Translation Lab Inc.}
\date{May 2026}

\begin{document}
\maketitle

\begin{abstract}
We present \textbf{Agentic AI Translate}, an agentic translator prototype that operationalises the thesis of Yamada (forthcoming) --- that the metalanguage of Translation Studies has become an instruction code for generative AI. The system replaces the dominant \emph{text-in / text-out} paradigm of machine translation with a four-stage agentic cycle (\textbf{Identify $\rightarrow$ Prompt $\rightarrow$ Generate $\rightarrow$ Verify}), preceded by an \textbf{interactive specification phase} in which the user composes --- through model-assisted dialogue --- a structured translation brief grounded in skopos theory, register, audience, and genre conventions. The verification stage adopts the GEMBA-MQM error-span protocol [Kocmi \& Federmann, 2023] for evidence-grounded scoring, and document-level coherence is preserved through a \emph{DelTA-lite} memory of proper nouns and a running bilingual summary, after [Wang et al., 2025]. We describe the philosophical motivation, the architectural commitments, the four reference-material categories the system consumes, and the principal design tensions the architecture makes explicit. Empirical validation is left for future work; the contribution here is conceptual and architectural --- an executable embodiment of the position that \emph{translation in the GenAI era is communication design, not text conversion}.
\end{abstract}


\section{Introduction}

For four decades, machine translation research has been organised around a single optimisation target: lexical and grammatical fidelity between a source string and a target string. Statistical and neural systems progressively closed the accuracy gap to professional human output across high-resource pairs, and large language models (LLMs) have now made fluency, idiomaticity, and basic register-matching nearly free at the segment level [Kocmi et al., 2024; Karpinska \& Iyyer, 2023]. In Kano-model terms [Kano, 1984], \textbf{accuracy has saturated as a Must-Be quality}: its presence no longer differentiates translations, only its absence is noticed.

The frontier of translation value has therefore moved to what Tannen (1986) called the \emph{how} rather than the \emph{what} --- register, audience design, voice, stance, cultural framing, genre convention --- the dimensions that have always mattered to professional translators but which computational research has historically left implicit. Yamada (forthcoming), in \emph{Metalanguage and GenAI: Empowering Language Learners and Translators in Training} (forthcoming in \emph{The Routledge Handbook of Translation and Technology}, 2nd ed.), argues that this is not merely a shift in evaluation criteria but a fundamental \textbf{reconfiguration of the translator's role}: from the manual drafter of target text toward the \textbf{designer of conditions} under which a generative system produces text --- and the \textbf{verifier} of whether that text fulfils its communicative purpose. Crucially, Yamada observes:

\begin{quote}\itshape "The easier it becomes to generate text, the harder it becomes to ensure that text fulfils a specific communicative purpose."\end{quote}

This \emph{automation paradox} dissolves once we recognise that the vocabulary of Translation Studies (TS) --- \emph{skopos, register, audience, equivalence, foreignization, domestication, genre, stance, loyalty} --- provides exactly the descriptive precision an LLM needs as instruction. \textbf{Theory becomes operational.} What was previously studied to think about practice is now spoken to instruct the machine.

This paper presents an executable embodiment of that argument. \textbf{Agentic AI Translate} is a research prototype and an agentic translator, openly released, that takes a translation request and walks the user through a structured authoring of a translation specification before any generation occurs. It then runs an agentic four-stage pipeline (\emph{Identification $\rightarrow$ Prompting $\rightarrow$ Generation $\rightarrow$ Verification}) that uses the specification end-to-end, with document-level state to preserve terminological consistency across long inputs. The contribution is not empirical --- we have not yet conducted the comparative MQM study against unstructured prompting that would validate the hypothesis --- but architectural: an executable description of \emph{what such a system must contain} if the position outlined above is to be realised in code.

The remainder of the paper is organised as follows. Section 2 develops the philosophical motivation. Section 3 specifies the architecture. Section 4 details the implementation. Section 5 positions the system relative to recent work in agentic LLM translation, document-level MT, and translation evaluation. Section 6 discusses limitations and the main design tensions. Section 7 outlines the validation plan and the structured-spec extension that constitutes the project's main research direction.

\section{Philosophical Motivation: Translation as Communication Design}

\subsection{The two layers, restated for the GenAI era}

Translation has always operated on two layers: the propositional content (the \emph{what}) and the realisation that content takes in the target language (the \emph{how}) --- including register, sentence rhythm, sociolectal markers, footnoting practice, the management of culturally bound items, and the addressivity that positions the implied reader. House's (2015) overt/covert distinction, Reiss's (1971/2000) text-typology, Nord's (1997) functionalist framing, and Vermeer's (1978) skopos all foreground the priority of communicative purpose over surface equivalence; this is not a new observation but a consensus in TS [see Munday, 2016, for the canonical synthesis].

The \emph{new} observation is that, until recently, encoding such constraints into a translation system meant either training a domain-specific model or post-editing the output of a generic one. Both approaches treated communicative design as something \emph{applied to} translation rather than something \emph{constitutive of} it. Generative LLMs change this: they accept long, structured natural-language instructions at inference time and condition their generation on those instructions to a degree that is qualitatively different from prior systems [Vilar et al., 2023; Karpinska \& Iyyer, 2023]. \textbf{Communicative design becomes a first-class input.}

\subsection{Voice as the unit of translation}

Consider Murakami Haruki's translation of Salinger's \emph{The Catcher in the Rye} into Japanese. The opening --- "If you really want to hear about it..." --- has been rendered by multiple Japanese translators with varying degrees of literal fidelity, but Murakami's choice deliberately preserves not the surface lexicon of Salinger but the \emph{voice} of Holden Caulfield: a particular cadence, a particular relation to the reader. A few-shot prompt to a current LLM, given a short Murakami sample, can reproduce that voice for adjacent passages with striking fidelity --- not because the LLM has read Salinger or Murakami, but because the voice has been \textbf{specified} as a constraint that the model can honour.

This is the operational core of the \emph{translation-as-design} claim: voice --- the most apparently artisanal aspect of literary translation --- turns out to be \textbf{specifiable}, and once specified, it is reproducible at scale. The translator's contribution is no longer manual drafting of every sentence, but the \textbf{design of voice as a constraint}.

\subsection{The translator's reconfiguration}

Yamada (forthcoming) frames the translator's emerging role as \textbf{designer + verifier}:

\begin{itemize}[leftmargin=*]
\item \textbf{Designer}: composes, with metalinguistic precision, the situational analysis (skopos, audience, register, genre) and the operational artefacts (glossaries, paired examples, parallel-text exemplars) that condition the generative system.
\item \textbf{Verifier}: judges output not as a post-edit task --- surface error correction --- but as a \emph{functional} and \emph{epistemic} judgement: does the output land for the audience? does it preserve factual structure? does it match the spec?
\end{itemize}

The critical pedagogical implication is that the \textbf{vocabulary} of TS --- what Gambier (2009) called the \emph{meta-language} of the discipline --- is no longer studied to \emph{think about} practice but to \emph{instruct} the machine. The discipline's theoretical apparatus becomes operational infrastructure. The system described here is built on that recognition.

\section{Architecture}

The system comprises three concentric layers: the \textbf{four-stage cycle} (the pipeline), the \textbf{interactive specification} that conditions every stage, and the \textbf{persistent state} that preserves document-level coherence.

\subsection{The four-stage cycle}

\begin{verbatim}
        +---------------------------------------------------------+
        |  1) Identification                                       |
        |     LLM extracts {skopos, audience, register,           |
        |     genre, stance, notes} as JSON from the source.      |
        +---------------------------------------------------------+
        |  2) Prompting                                            |
        |     Deterministic Python composes a translation prompt  |
        |     from spec + references + identification + memory.   |
        +---------------------------------------------------------+
        |  3) Generation                                           |
        |     Single LLM call produces the draft (T = 0.3).       |
        +---------------------------------------------------------+
        |  4) Verification                                         |
        |     LLM-as-judge returns MQM error spans                |
        |     {span, category, severity, explanation}.            |
        |     Score = -25*crit -5*major -1*minor.                 |
        |     Verdict computed deterministically vs. threshold.   |
        |     If revise: errors fed back as Stage 2 refinement;   |
        |     up to two iterations.                               |
        +---------------------------------------------------------+
\end{verbatim}

\textbf{Why four stages, not one?} A single end-to-end prompt forces the model to perform situational analysis, prompt assembly, generation, and self-evaluation in a single forward pass --- producing fluent but largely unanalysable output. Decomposition exposes each commitment as an inspectable artefact. The Identification JSON, the assembled Stage 2 prompt, and the Verification error spans are all logged and visualised in the user interface; this is by design, since the pedagogical and research value of the system depends on each stage being legible.

\textbf{Why Stage 1 is a separate LLM call.} Situational analysis can in principle be folded into a single generation prompt. We separate it for two reasons. First, the JSON it produces --- \texttt{\{skopos, audience, register, genre, stance, notes\}} --- is the \emph{most direct embodiment} of the metalanguage thesis: TS categories appear as structured fields, not as prose. Second, separating it allows the user to \emph{see} the situational analysis the model has performed and challenge it before generation. In current practice this remains a read-only artefact; making it user-editable is a planned extension.

\subsection{Interactive specification}

The most distinctive element of the system is the layer that precedes the pipeline. After source text entry, the user clicks \textbf{Propose spec}; the model returns a structured markdown document with ten canonical sections --- \emph{Skopos, Audience, Register \& Voice, Genre, Terminology guidance, Style decisions, Things to preserve, Things to localise, Things to avoid, Open questions} --- drafted from the source and any uploaded references. The user may:

\begin{enumerate}[leftmargin=*]
\item Edit the markdown directly in the UI;
\item Refine via chat (``audience is academic peer reviewers'', ``use plain da/dearu style throughout'', ``preserve emoji and source-language fan vocabulary'');
\item Iterate until satisfied, then \textbf{lock} the spec, after which translation may run.
\end{enumerate}

The lock step is intentional. It enforces an \textbf{architectural commitment} that no translation can be produced without an explicit, user-endorsed specification. The system therefore cannot be used as a generic MT tool; it can only be used as a \emph{spec-driven translation tool}. This is the philosophical position made operational.

The specification is consumed identically by Stage 2 (Prompting) and Stage 4 (Verification): the verifier judges the translation against the same spec the generator was conditioned on. This closes a common evaluation loophole in which the verifier and the generator implicitly disagree about what counts as good output.

\subsection{Reference-material layer}

Four orthogonal categories of reference materials may be uploaded:

\begin{center}
\begin{tabular}{p{3.2cm} p{4.2cm} p{5.2cm}}
\hline
\textbf{Category} & \textbf{Format} & \textbf{Functional role} \\
\hline
Glossary & TSV/CSV (source $\leftrightarrow$ target) & Mandatory terminology \\
Paired examples & TSV/CSV (source $\leftrightarrow$ target) & Translation-judgement few-shot \\
Parallel target-language texts & TXT/MD & Genre-voice exemplars \\
Style guide & MD/TXT & Free-form narrative constraint \\
\hline
\end{tabular}
\end{center}

These categories follow the pragmatic taxonomy used in professional CAT/TMS workflows and partially align with the ASTM F2575 standard for translation specifications. The system injects all four into the spec proposal, the generation prompt, and the verifier --- each consumer can decide how to weigh them. The current implementation injects all paired examples; selective retrieval (R-BM25 or embedding similarity, after Agrawal et al., 2023) is a planned upgrade.

\subsection{Document-level memory (DelTA-lite)}

For multi-paragraph inputs, the chunker splits the document at blank-line paragraph boundaries (with sentence-boundary fallback for over-long paragraphs). Each chunk is translated independently, but between chunks a \textbf{persistent memory} is updated by an auxiliary LLM call, after Wang et al. (2025):

\begin{itemize}[leftmargin=*]
\item \textbf{Proper-noun ledger}: a running source-to-target dictionary of terms whose translations should remain stable (people, places, organisations, products, technical terms).
\item \textbf{Bilingual running summary}: 50--150 words in the target language, capturing the document's progression for tonal continuity.
\item \textbf{Immediate-window context}: the previous chunk's source and target text.
\end{itemize}

These three artefacts are injected into the next chunk's Stage 2 prompt under explicit headings (\emph{Established terminology}, \emph{Document summary so far}, \emph{Immediately preceding chunk}) and the model is instructed to honour them. In informal observation on multi-paragraph literary and journalistic test inputs, the ledger correctly captures named entities that recur across chunks (e.g., \emph{Natsume Soseki $\rightarrow$ Natsume Soseki}, \emph{Kushami-sensei $\rightarrow$ Kushami}) without further intervention, mirroring the consistency improvements reported by Wang et al. (2025) at scale.

\subsection{MQM-grounded verification}

Stage 4 follows the \textbf{GEMBA-MQM} protocol of Kocmi \& Federmann (2023): the verifier prompt is language-agnostic, instructs the model to identify error spans and assign each one an MQM category and severity, and returns a structured JSON list. The category set is the canonical inventory of Freitag et al. (2021): \emph{Accuracy} (mistranslation, addition, omission, untranslated, do-not-translate), \emph{Fluency} (grammar, punctuation, spelling, register, inconsistency, character encoding), \emph{Terminology, Style, Locale convention, Other}. Severity is one of \emph{critical, major, minor}. From the error list a deterministic score is computed:

\[
\mathrm{score} = -25 \cdot n_{\text{critical}} - 5 \cdot n_{\text{major}} - 1 \cdot n_{\text{minor}}
\]

The verdict is \emph{accept} if the score meets a configurable threshold (default -2, i.e. up to two minor issues are tolerated; any major or critical triggers revision), otherwise \emph{revise}. On revision, the typed error list is appended verbatim to the Stage 2 prompt as actionable instructions, and Stage 3 re-runs. The loop is bounded at two iterations --- both Huang et al. (2024) and Stechly et al. (2024) show that intrinsic LLM self-correction yields rapidly diminishing returns and can degrade output.

We follow Fernandes et al. (2023) and Wang et al. (2024) in requiring the verifier to \textbf{emit evidence (error spans) before the score}, which empirically reduces verbosity and self-preference biases in LLM-as-judge configurations.

\section{Implementation}

The system is implemented in approximately 1200 lines of Python (excluding prompts and tests). The runtime stack is:

\begin{itemize}[leftmargin=*]
\item \textbf{Anthropic SDK} with Claude Sonnet 4.6 as the default model (configurable);
\item \textbf{Streamlit} for the UI;
\item \textbf{python-dotenv} for local development; in deployment the API key is supplied per-session by the user via the sidebar (no shared key);
\item No vector database, no GPU, no fine-tuning.
\end{itemize}

The minimal stack is intentional: every commitment in the system is in \emph{prompts} and \emph{Python flow control}, not in trained weights. This makes the system fully inspectable and reproducible, and keeps the cost of experimenting with alternative spec structures or verifier prompts at the level of editing a text file.

The repository is publicly available on GitHub under MIT licence (\textcopyright{} Translation Lab Inc.) at \url{https://github.com/chuckmy/agentic-translator}, and a live demo is deployed on Streamlit Community Cloud at \url{https://agentic-translator-chuckmy.streamlit.app}, with bring-your-own API key. A bilingual test set covering three genres (news, literary description, academic abstract) in both translation directions is included to support reproducible exploration.

\section{Related Work}

\textbf{Spec-aware MT.} The closest precursor is Kayano \& Sugawara (2025), who demonstrate that prompting with an explicit translation specification --- purpose, audience, register --- significantly improves preference scores on intent-rich texts, sometimes exceeding human reference translations. Their specification is presented as flat free-form text; we extend the idea by making the specification an interactively \emph{authored} artefact with a stable structural template, and by carrying it through both generation and verification rather than only generation.

\textbf{Multi-agent translation.} Wu et al. (2024/2025) introduce \textbf{TransAgents}, a six-agent simulation (CEO, editor, translator, localizer, proofreader, QA) for ultra-long literary texts; the system is preferred by both expert and crowd judges over GPT-4 single-call output and over reference human translations on book-length input, despite \emph{lower} d-BLEU. This is the strongest available evidence that pipeline decomposition is qualitatively beneficial for the \emph{attractive-quality} dimensions identified in \S{}2. Briakou et al. (2024)'s \textbf{Translating Step-by-Step} demonstrates the same principle within a single model --- pre-translation research $\rightarrow$ drafting $\rightarrow$ refining $\rightarrow$ proofreading --- establishing WMT24 SOTA. Our prototype is closer in shape to the latter than to TransAgents; planned extensions (R5, \S{}7) move toward role decomposition.

\textbf{Document-level translation.} Karpinska \& Iyyer (2023) document the strong human-evaluation preference for paragraph-level over sentence-level LLM translation, especially in literary registers. \textbf{DelTA} (Wang et al., 2025) introduces explicit four-tier memory --- proper nouns, bilingual summary, long-term, short-term --- with measurable consistency gains. Our DelTA-lite implements the first two tiers.

\textbf{LLM-as-judge for translation.} Kocmi \& Federmann (2023) establish that GPT-4 with a fixed three-shot MQM prompt produces scores correlating with expert MQM well enough to win the WMT23 metrics task. xCOMET (Guerreiro et al., 2024) and MetricX (Juraska et al., 2023) demonstrate that \emph{learned} metrics still outperform LLM-judge prompts at the segment level on WMT24/25; we accept this and treat the LLM-judge as the system's first line of evaluation while documenting (\S{}6) the planned augmentation by xCOMET as an external signal.

\textbf{Self-correction in MT.} Madaan et al. (2023) introduce Self-Refine; Feng et al. (2025)'s \textbf{TEaR} shows MQM-typed feedback improves refinement quality. Huang et al. (2024) and Stechly et al. (2024) document the limits of intrinsic self-correction --- apparent improvements are often artefacts of sampling diversity rather than genuine self-critique. We bound our revise loop accordingly.

\textbf{Translation Studies frameworks operationalised.} Singh et al. (2024) explore cultural and register-specific adaptation with LLMs, and a growing body of work investigates honorific and politeness-marker handling in LLM-mediated translation. The systematic encoding of TS frameworks (Reiss, Nord, House) as machine-readable schema remains an open area, identified in \S{}7 as our primary research direction.

\section{Discussion and Limitations}

\subsection{What the prototype claims, and what it does not}

This is an \textbf{architectural} contribution. We claim that the architecture coherently embodies the \emph{translation-as-design} position; we do not yet claim that it produces measurably better translations than alternatives. A controlled comparison --- same source, same target, same model, with vs. without spec --- across multiple genres and language pairs, evaluated by professional translators using full MQM, is the necessary next step.

\subsection{Single-model verification is the weakest link}

The verifier currently runs on the same model family as the generator. This exposes the system to \textbf{self-preference bias} [Zheng et al., 2023; Wang et al., 2024] and to the broader limits of intrinsic self-correction [Huang et al., 2024; Stechly et al., 2024]. The bounded loop (two iterations) and the evidence-first prompt structure (errors before scores) mitigate but do not resolve this. The planned augmentation (R2, \S{}7) introduces a cross-model judge and an external learned QE signal (xCOMET-XL or MetricX).

\subsection{The spec is currently free-form markdown}

The interactive specification is a markdown document with ten canonical headings but otherwise unconstrained content. This permits expressive richness but limits machine readability and forecloses systematic A/B experimentation on individual fields (e.g., does specifying \emph{loyalty target} produce measurable behavioural change?). The planned extension (R6, \S{}7) replaces the markdown with a structured JSON schema operationalising Reiss's text typology, Nord's loyalty, House's overt/covert mode, and a domestication--foreignization continuum, with the markdown view derived from the schema. This is the project's primary research direction.

\subsection{Reference materials are injected without selection}

All paired examples are currently injected into every prompt. Agrawal et al. (2023) show that even one mismatched in-context example can degrade translation quality more than no examples at all; Vilar et al. (2023) show that example \emph{quality} dominates over similarity at large model scale. Both effects motivate retrieval-based selection (R-BM25 plus embedding similarity over a quality-tagged TU store), which is currently absent.

\subsection{No empirical evaluation}

The most important limitation. The validation plan is sketched in \S{}7.1.

\section{Future Work}

\subsection{Empirical validation}

A factorial study comparing (a) generic-prompt translation, (b) free-form spec, (c) structured-schema spec, across (i) literary, (ii) journalistic, (iii) academic genres, in ($\alpha$) JA$\rightarrow$EN and ($\beta$) EN$\rightarrow$JA, evaluated by professional translators using Freitag-2021 MQM with ESA severity protocol, would establish whether and where the spec-driven architecture yields measurable gains. Inter-rater agreement and per-axis breakdown (accuracy vs. style vs. terminology) would identify which dimensions of \emph{attractive quality} are most spec-sensitive.

\subsection{Structured specification schema (R6)}

Replace the markdown spec with a JSON schema:

\begin{verbatim}
{
  "skopos": "...",
  "text_type": "informative | expressive | operative | audiomedial",
  "house_mode": "overt | covert",
  "loyalty": { "author_intention": 0.7,
               "ST_culture_fidelity": 0.5,
               "TT_reader_orientation": 0.9,
               "commissioner_brief": 0.6 },
  "domestication_axis": 0.7,
  "audience": { ... },
  "register": { ... },
  "preserve": [...], "localize": [...], "avoid": [...]
}
\end{verbatim}

Reiss's text typology, Nord's loyalty principle, House's overt/covert distinction, and the Schleiermacher--Venuti domestication--foreignization axis become \textbf{fields the user fills in}. The same schema underwrites the generation prompt, the verification prompt, and (critically) the experimental design --- fields can be ablated, swapped, or held constant across runs to support A/B research that is structurally impossible with free-form spec.

\subsection{Multi-agent decomposition}

Following TransAgents and Briakou et al., split Stage 3 into \textbf{research $\rightarrow$ draft $\rightarrow$ localise $\rightarrow$ proofread}, with each pass governed by the same locked specification. Particular value is expected in the \emph{localise} pass (cultural rendering, idiom handling), which TransAgents identified as the largest contributor to expert preference.

\subsection{External quality signals}

Add \textbf{xCOMET-XL} or \textbf{MetricX} (Juraska et al., 2023) as a parallel verifier whose score gates acceptance alongside the LLM judge. Cross-model judging (Sonnet generator + Gemini or GPT-5 judge) addresses self-preference bias.

\subsection{Hallucination check}

Augment Stage 4 with an entity-preservation check (named entities and numerals in source must appear or have explicit equivalents in target), per Guerreiro et al. (2023). Yamada's \emph{factual} axis is currently the weakest detector for fluent-but-fabricated output.

\section{Conclusion}

We have described a research prototype that embodies, in executable form, the position that translation in the GenAI era is \emph{communication design}. The core architectural commitments --- interactive specification, the four-stage cycle, document-level memory, MQM-grounded evidence-first verification --- are not arbitrary engineering choices but direct operationalisations of the metalanguage thesis advanced by Yamada (forthcoming). The system is openly released so that colleagues, students, and researchers can interrogate, extend, and contest the position it makes operational.

What remains is the empirical work that this paper has explicitly not undertaken: the controlled study that would tell us whether \emph{making the specification explicit} produces, in measurable MQM terms, the qualitative shift in \emph{attractive quality} that the lecture-platform argument predicts. That study, and the structured-schema extension that it requires, constitutes the project's next phase.

\end{document}